\documentclass[5p,times]{elsarticle}
\usepackage[T1]{fontenc}
\usepackage[utf8]{inputenc}
\usepackage{courier,graphicx,amsmath,amssymb,booktabs,xcolor,microtype,xurl,hyperref,placeins,capt-of,etoolbox,dblfloatfix,balance}
\definecolor{WasserLink}{HTML}{163E70}
\definecolor{WasserURL}{HTML}{0F766E}
\hypersetup{colorlinks=true,linkcolor=WasserLink,citecolor=WasserLink,urlcolor=WasserURL,
  pdftitle={WasserMan: Benchmark for Underwater Manipulation Policy Learning},
  pdfauthor={Danil Belov, Artem Erkhov, Sergei Parsegov, Pavel Osinenko},
  pdfsubject={Simulation benchmark with paired visual, physical and execution diagnostics}}
\AtBeginDocument{\hypersetup{linkcolor=WasserLink,citecolor=WasserLink,urlcolor=WasserURL}}
\myfooter[L]{Preprint\hfill 20 October 2026}
\newcommand{\WasserManResources}{%
  {\footnotesize
  \href{https://dancher00.github.io/wasserman/}{Website}\quad$\cdot$\quad
  \href{https://github.com/dancher00/wasserman}{GitHub}\quad$\cdot$\quad
  \href{https://huggingface.co/datasets/dancher00/WasserMan}{Hugging Face}\par}
  \vskip7pt
}
\newcommand{\WasserManTeaser}{%
  \begin{minipage}{\textwidth}
  \normalfont
  \centering
  \WasserManResources
  \includegraphics[width=\textwidth]{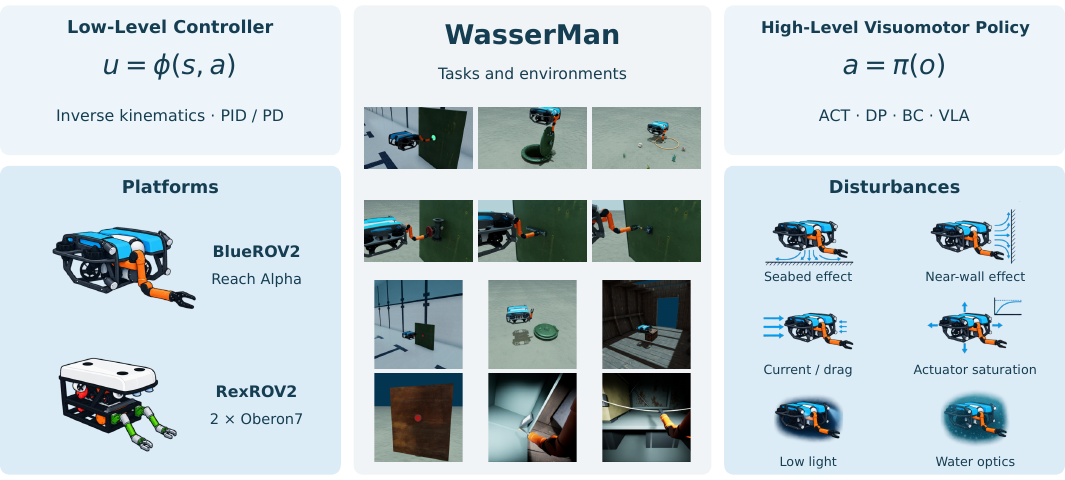}
  \captionof{figure}{WasserMan overview: execution controllers and vehicle--arm platforms (left), contact tasks and simulated environments (center), visuomotor policies, physical disturbances and visual conditions (right).}
  \label{fig:teaser}
  \end{minipage}\par\vskip12pt
}
\makeatletter
\patchcmd{\MaketitleBox}{\vskip36pt}{\vskip18pt}{}{\errmessage{Title spacing hook unavailable}}
\patchcmd{\MaketitleBox}{\hrule\vskip12pt}{\WasserManTeaser\hrule\vskip12pt}{}{\errmessage{Opening teaser hook unavailable}}
\makeatother
\begin{document}
\begin{frontmatter}
\title{WasserMan: Benchmark for Underwater Manipulation Policy Learning}
\author[cu,cess,sirius]{Danil Belov\corref{corresponding}}
\ead{dancher.00@gmail.com}
\author[cess]{Artem Erkhov}
\author[sirius,ics]{Sergei Parsegov}
\author[cu,cess,sirius]{Pavel Osinenko}
\cortext[corresponding]{Corresponding author.}
\address[cu]{Central University, Moscow, Russia}
\address[cess]{Center for Engineering Systems and Sciences, Moscow, Russia}
\address[sirius]{Sirius University of Science and Technology, Sirius Federal Territory, Russia}
\address[ics]{V.~A.~Trapeznikov Institute of Control Sciences of Russian Academy of Sciences, Moscow, Russia}

\begin{abstract}
\textbf{Purpose:} WasserMan provides a reproducible simulation benchmark for training and evaluating underwater visuomotor manipulation policies, whose execution couples contact, vehicle stabilization and actuator limits.
\textbf{Methods:} Six tasks combine fixed demonstration splits and training budgets with full-horizon physical success criteria, reset-matched expert references and common-exposure diagnostics that retain failed trials. Reference implementations compare ACT, diffusion policies and behavioral cloning across six tasks, with an additional three-task pretrained SmolVLA evaluation. Each uses three training runs per task; paired interventions hold policy weights fixed.
\textbf{Results:} Across 54 primary training runs and 1,620 evaluated episodes, the benchmark establishes native performance. Interventions reveal three limitations of success-only evaluation: policies fail in expert-solvable scenes, increased early opposing grasp coexists with failed completion, and preserved completion conceals increased actuator demand. RGB controls distinguish image freshness from correspondence to the physical reset.
\textbf{Conclusions:} The benchmark and paired protocol support reproducible learning comparisons and execution diagnostics beyond success rates. Versioned code, demonstrations, checkpoints and physical traces are released. The evidence is simulation-only; underwater image realism and transfer to hardware remain unvalidated.
\end{abstract}
\end{frontmatter}
\raggedbottom
\section{Introduction}
Underwater manipulation couples visual decisions, arm contact and vehicle
stabilization~\cite{fossen,girona,carrera,simetti2015}. A policy must reach a
mechanism, maintain useful contact and complete its motion while the vehicle
responds to fluid forces and actuator limits. A binary success rate records
whether this sequence ends correctly. It does not reveal whether a changed
scene remains physically solvable, whether additional contact advances the
task, or whether successful execution requires greater actuator demand.
These distinctions matter when comparing a policy before and after a change
in its observations, controller or environment.

WasserMan provides a simulation benchmark for training and evaluating
visuomotor policies on six underwater manipulation tasks. It specifies task
contracts, demonstration data and whole-episode splits, training exposure and
test resets, and releases reference implementations, checkpoints and
per-episode physical traces. A paired evaluation protocol connects task
completion to execution. The protocol holds policy
weights fixed, matches initial states between an intervention and its native
reference, and scores the same full-horizon physical task contract. An
unchanged privileged expert is evaluated on the same resets for physical-scene
interventions. Contact, mechanism progress, tracking and actuator measurements
are compared over a common active exposure, retaining failed episodes. Thus a
failure remains available for physical comparison rather than disappearing
from a success-conditioned diagnostic summary.

Three experiments illustrate the resulting scientific account. Translating
OpenHatch by $\pm4$\,cm eliminates DP completion while the expert continues to
complete every matched reset. Removing the controller's integral action also
eliminates completion, but increases early opposing-grasp occupancy. Applying
current preserves hatch completion while increasing motor-force RMS. These
are different outcomes of a coupled robot--policy system: failure in an
expert-solvable scene, increased contact without completion, and increased
actuator demand behind an unchanged success rate. Each requires evidence that
completion alone does not supply.

We make two contributions. First, the six-task benchmark standardizes
demonstration-based learning and physical evaluation, with reference Action
Chunking with Transformers (ACT), diffusion-policy (DP) and behavioral-cloning
(BC) implementations, three training runs per configuration and versioned
reproduction records. Second, the paired protocol combines full-horizon task
contracts, reset-matched expert references and common-exposure diagnostics
that retain failures. The three cases above demonstrate its scientific value;
RGB controls additionally separate changing visual feedback from its
correspondence to the current physical reset.

The paper follows these questions rather than the chronology of the
experimental campaigns. After introducing the benchmark and protocol, we
report reference performance, examine the three diagnostic cases and test
visual feedback. Additional platforms, action interfaces, training ablations,
candidate-stream audits and complete reproduction specifications are retained
in the Supplementary Materials. The opening page links the project website,
GitHub repository and Hugging Face release; Table~\ref{tab:release-contract}
identifies the code, demonstrations, policy models and evaluation records.

\section{Related Work}
\textbf{Underwater simulation and learning.}
UUV Simulator, DAVE and Stonefish provide marine dynamics, sensors and
intervention capabilities~\cite{uuv,dave,stonefish,stonefishml}. DAVE includes
stabilized bimanual manipulation; URoBench compares simulation tools for
reinforcement learning~\cite{urobench}, and MarineGym supports GPU-parallel
vehicle-control tasks~\cite{marinegym}. Earlier intervention benchmarks address
tracking and reconstruction~\cite{sanz2015benchmark}. OceanSim compares its
perception simulation with real camera and sonar data~\cite{oceansim}.
These systems establish important simulation and evaluation capabilities.
WasserMan focuses on the relation between manipulation completion, expert
solvability and the physical execution of a fixed policy.

Underwater learning systems cover complementary task and data settings.
USIM/U0 provides multi-task simulation data and task-specific completion
criteria for VLA evaluation~\cite{usim}; AquaCap evaluates generated-code
policies in simulation and on an ROV~\cite{aquacap}. AquaBot studies
self-improving manipulation on hardware~\cite{aquabot}, while Bi-AQUA,
UMI-Underwater and ULOHA investigate imitation, demonstration transfer and
bimanual learning~\cite{biaqua,umiunderwater,uloha}. Here the reference
implementations use fixed demonstration splits and sample budgets, followed
by interventions that relate their completion outcomes to recorded execution.

\textbf{System-level evaluation.}
AM-Bench is the closest system-level comparator~\cite{ambench}. It evaluates
aerial contact and transport jointly with interfaces, embodiments, disturbances
and actuator constraints. Its reported tracking and limit-use diagnostics
average successful rollouts. WasserMan combines underwater mechanism tasks
with expert references on intervention-matched resets and physical summaries
that retain unsuccessful rollouts, including conditions with no completions.
This makes it possible to compare contact and actuator behavior precisely
where a success-conditioned summary would have no sample. Supplementary
Table~S11 retains the source-checked feature matrix and its
scope definitions.

\textbf{Manipulation benchmarks and reference methods.}
RLBench, robosuite and ManiSkill2 provide broader manipulation
suites~\cite{rlbench,robosuite,maniskill2}. Robomimic motivates careful
comparisons of complete imitation-learning implementations~\cite{robomimic}.
Our reference policies combine established ACT and diffusion methods, visual
encoders and sampling procedures~\cite{act,dp,resnet,clip,ddim} on Isaac
Lab~\cite{isaaclab}. Underwater image formation, illumination and refraction
motivate controlled visual changes~\cite{akkaynak,seathru,illumination,refractive,calibration};
dynamics randomization motivates testing parameter sensitivity~\cite{peng}.
The question addressed here is how these controlled changes alter solvability,
contact and actuator demand under a shared evaluation contract.

\section{WasserMan Benchmark}
\label{sec:benchmark}
\subsection{Tasks and physical contracts}
\begin{figure*}[tbp]
\centering
\includegraphics[width=\textwidth]{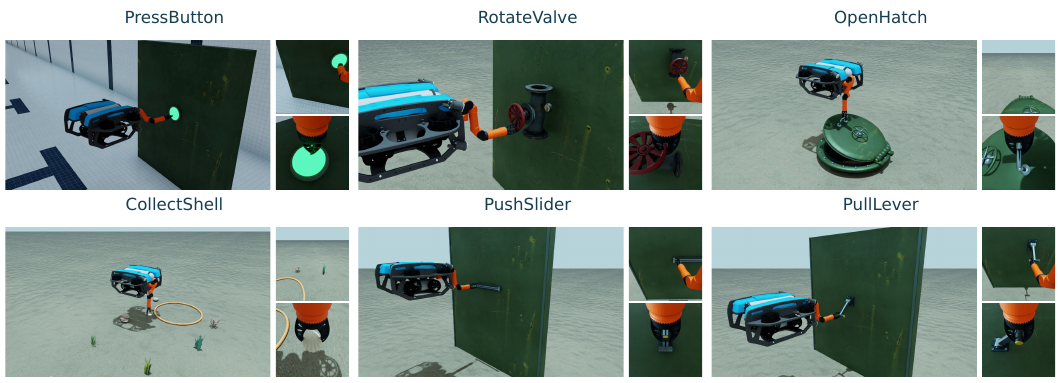}
\caption{Six learning tasks, shown at one expert state per task. The external,
base and wrist views are synchronized. These CAD-profile views illustrate
the mechanisms; policy learning uses procedural visuals and box fingers.}
\label{fig:open-tasks}
\end{figure*}

The benchmark contains six learning tasks: PressButton, RotateValve,
OpenHatch, CollectShell, PushSlider and PullLever (Fig.~\ref{fig:open-tasks}).
They span stable pressing, sustained rotation, grasped opening, transport and
release, and engagement with translating or rotating mechanisms. A task
contract fixes its observations, command decoding, reset generator, episode
horizon and physical completion predicate. Completion is independently
rescored from the physical trace rather than inferred from rendered motion
or from a policy's command.

PressButton requires 4--7\,mm of travel and a continuous one-second stable
contact hold. RotateValve requires $170^\circ$ of signed rotation.
OpenHatch requires an opening in the $80$--$105^\circ$ interval, at least
$75^\circ$ of motion under an opposing grasp, at most $5^\circ$ of ungrasped
motion and a continuous one-second hold satisfying the speed and stabilization
tests. CollectShell requires grasping, transport, full-footprint placement,
release and settling. PushSlider requires 0.27\,m of travel and PullLever
$40^\circ$ of rotation, both with engagement and four stable control steps.
These conjunctions distinguish incidental contact or partial progress from
a completed manipulation.

The contracts also determine which physical records are informative for each
task. Button travel must be accompanied by stable contact; valve rotation
has a signed direction; hatch opening must occur predominantly under grasp.
Shell transport additionally needs a valid placement and release. For the
slider and lever, engagement and stabilization matter alongside displacement.
Consequently, an evaluator can retain intermediate measurements while keeping
one unchanged binary definition of completion. A policy that produces motion
or contact without satisfying the relevant conjunction remains a failed
execution, and its trace records the unsatisfied conditions. This provides a
common evaluation principle across the six tasks without equating the
mechanisms' distinct physical requirements.

\begin{table}[tbp]
\centering
\caption{Task command contracts. $d_u$: decoded command size; $f_\pi$: target
rate; $H$: full horizon in 30\,Hz control steps. Every policy predicts 16
targets and executes eight.}
\label{tab:open-contracts}
\small
\begin{tabular}{llrrr}
\toprule
Task & Interface & $d_u$ & $f_\pi$ (Hz) & $H$ \\
\midrule
PressButton & Actuator & 10 & 30 & 470 \\
RotateValve & EE pose/jaw & 8 & 30 & 2240 \\
OpenHatch & Actuator & 8 & 20 & 1790 \\
CollectShell & Actuator & 8 & 30 & 7190 \\
PushSlider & Actuator & 8 & 30 & 1790 \\
PullLever & Actuator & 8 & 30 & 1790 \\
\bottomrule
\end{tabular}
\end{table}

\subsection{Observations, expert demonstrations and execution}
Policies receive wrist RGB and measured robot state. Object state, expert
phase and contact evidence are excluded from policy inputs. This separation
allows the evaluator to use physical measurements without supplying them to
the learned controller. A state-PPO expert supplies button
demonstrations~\cite{ppo}; scripted feedback supplies the other tasks. The
expert has privileged information and serves as a reference execution, not as
a learned visual baseline.

The \texttt{open-procedural-v1} learning profile uses BlueROV2/Reach Alpha,
primitive arm visuals and two-segment box fingers, with the versioned joint
frames, inertias and remaining collision proxies retained. The release also
supports RexROV2/Oberon7 and a twin-arm
configuration. Their separate CAD-profile expert studies are documented in
Supplementary Section~S2; they are not learned-policy
transfer experiments.

End-effector targets pass through inverse kinematics~\cite{nakamura}; native
base/joint targets bypass that conversion. Vehicle feedback and joint servos
drive separate force and torque paths. Thruster allocation is bounded by the
manufacturer's T200 16\,V static thrust curve~\cite{t200}, followed by actuator
lag and delay. The reduced-order hydrodynamics combine buoyancy, diagonal
added mass, Coriolis terms and linear/quadratic damping~\cite{fossen}, using
link velocity relative to the water. Contact, tracking and applied motor
forces are recorded from this coupled execution. Supplementary
Fig.~S1 gives the full execution diagram and parameter provenance.

\subsection{Water effects in visual observations}
\begin{figure}[tbp]
\centering
\includegraphics[width=\columnwidth]{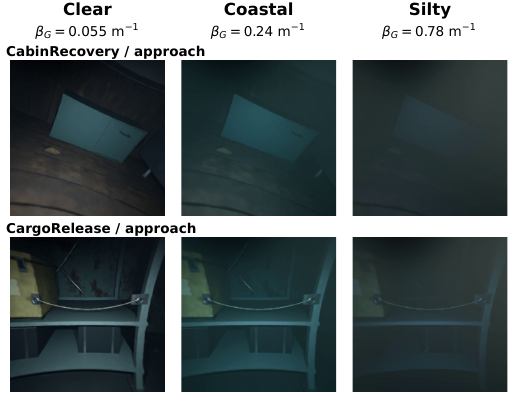}
\caption{Water effects in matched, unretouched optical replays before close
approach. Camera, state and lighting are fixed within each row.
$\beta_G$: green-channel attenuation; backscatter and particles also vary.
These base-camera views illustrate the renderer's visibility controls;
learned policies receive wrist RGB.}
\label{fig:optical}
\end{figure}

Water appearance determines the RGB information available to a visuomotor
policy. The renderer supports flat-port refraction, wavelength-dependent
attenuation, backscatter, suspended sediment and robot-mounted lights.
Attenuation changes color and contrast with viewing distance; backscatter
adds a veiling component, while particles and lighting alter the visible
scene. Figure~\ref{fig:optical} shows clear, coastal and silty presets applied
to recorded expert trajectories. Camera, physical state and lighting are
matched within each row, exposing appearance changes in the same scene.

The policy tests in Section~\ref{sec:visual} use the actual wrist observations.
Clear, coastal and harbor profiles apply distance-dependent attenuation and
backscatter using wrist RGB and camera range. Separate dim and bright
conditions change scene lighting; particle and refractive settings remain
fixed in these tests. Supplementary Fig.~S4
shows the resulting policy inputs on one shared initial scene. The profiles
are synthetic conditions, not fits to measured water environments. Across
appearance interventions, policy weights, resets and physical task contracts
remain unchanged, allowing visual sensitivity to be evaluated separately
from interventions on the forces that the robot must counteract.

\section{Evaluation Protocol}
\label{sec:evaluation-design}
The protocol separates the final outcome from the evidence needed to interpret
it. For each intervention, we retain its own native reference, match the
initial robot, mechanism and fluid state, and keep the policy weights fixed.
Both executions run to the original horizon or their normal termination.
Full-horizon physical completion is reported alongside contact, progress,
tracking and actuator summaries over their common active exposure. Episodes
that fail the task remain in both the completion denominator and the physical
comparison.

The six-task benchmark fixes its protocol and asset profile.
Table~\ref{tab:release-contract} identifies these and the public distribution
snapshots for restoring its code, data and models.
Original run manifests retain the frozen scientific runtime, checkpoint and
cohort identities. A new policy is evaluated against the same task contracts,
demonstration split, sample budget and held-out reset protocol.

\begin{table}[tbp]
\centering
\caption{Six-task benchmark release identity. Linked revisions identify
immutable public distribution snapshots checked on 20 October 2026;
original execution provenance is retained in the run manifests.}
\label{tab:release-contract}
\small\setlength{\tabcolsep}{3pt}
\begin{tabular}{@{}p{0.32\columnwidth}p{0.64\columnwidth}@{}}
\toprule
Component & Identifier \\
\midrule
Protocol & \texttt{wm-open-v2-20260929} \\
Assets & \texttt{open-procedural-v1} \\
Code & \href{https://github.com/dancher00/wasserman/commit/283319443a07a740449d26260de2ac9d2ae75d0c}{\texttt{283319443a07}} \\
Demonstrations & \href{https://huggingface.co/datasets/dancher00/WasserMan/tree/00d9d5b191384d28111727d817c68d0ebbde802a}{\texttt{00d9d5b19138}} \\
Policy models & \href{https://huggingface.co/dancher00/WasserMan-models/tree/85de7d90e0e0eb2bdca194f40401a580bf7b263a}{\texttt{85de7d90e0e0}} \\
Study records & \href{https://huggingface.co/datasets/dancher00/WasserMan-Review-v3/tree/af9968b60e8bf534289a6653ae708fccb94dbf66}{\texttt{af9968b60e8b}} \\
\bottomrule
\end{tabular}
\end{table}

\subsection{Reference training and reset pairing}
Each task supplies the first 80 successful demonstrations from its configured
expert candidate stream. A fixed whole-episode permutation assigns 76 to
training and four to validation. All attempts, including failures and attempts
beyond the selected demonstrations, remain available for audit. The primary
test uses 30 fresh, unfiltered draws from the task's reset generator. Mechanism
position is fixed during valve and hatch collection; their translated
conditions therefore test placement transfer outside that training pose.

ACT, DP and BC use training seeds 17, 43 and 101 and equal exposure to
320,000 non-padded demonstration windows per run. ACT and BC use 10,000
updates with batch 32; DP uses 5,000 with effective batch 64. The final
checkpoint is evaluated without task-dependent checkpoint selection. ACT
uses ResNet18 and a transformer, DP uses CLIP ViT-B/16 and a conditional UNet,
and BC uses ResNet18 and a two-layer prediction head. ACT and BC observe one
image; DP observes two. All predict 16 targets and execute eight.
These are complete implementations: encoders, histories, representations and
compute differ, so the reference comparison does not isolate an algorithm
effect. Optimizers, schedules, precision, decoding and training costs are
specified in Supplementary Section~S1.

Within an intervention study, every condition uses the same ordered 30 resets
for each of the three checkpoints. Pre-warmup robot poses, joint states and
fluid draws match; the subsequent observations and actions may differ as a
consequence of the intervention. Separate studies use separate reset blocks
and new matched native references. For example, a primary-table rate is not
the baseline for a current experiment on another cohort. Supplementary
Table~S10 maps claims to their checkpoints, reset blocks and
asset profiles, and Supplementary Table~S1 records the
full reset distributions.

\subsection{Physical interventions and the expert reference}
The scene study changes one factor at a time on RotateValve and OpenHatch:
contact friction, actuator delay, mechanism damping, arm mass and inertia,
mechanism translation or assembly yaw. It evaluates each fixed DP checkpoint
and an unchanged privileged expert on the same reset identities. The
supporting geometry moves with a translated or rotated mechanism. Simulator
readback records verify the realized intervention settings. An expert
completion witnesses a completing execution under that task contract; expert
failure leaves feasibility unresolved. The expert is not assumed to be an
upper bound on the policy.
Expert completions are independently verified from their physical traces.

Controller interventions vary only the integral multiplier, holding the
policy weights, P/D gains, command interface and task predicate constant.
Additional coefficient replacements use published BlueROV2
parameters~\cite{vonbenzon2022}. Current interventions apply a world-frame
flow before camera warmup, with matching pre-current states and unchanged
gains, allocation limits and rendering. These changes operate through the
closed loop: fixed weights do not imply fixed commands, because the policy
continues receiving the resulting robot state and images.

\subsection{Common-exposure physical diagnostics}
Continuous diagnostics use the predeclared first 20\,s, ending at the earlier
active termination in each pair. Reset, warmup and inactive samples are
excluded, and exposure is recorded explicitly. For the principal comparisons
reported below, every reset pair supplies all 600 samples at the 30\,Hz control
rate. Completion still uses the complete task horizon in
Table~\ref{tab:open-contracts}. An early observation window therefore cannot
replace the completion predicate.

Let $\mathcal{T}_{c,s,r}$ contain the samples shared by intervention $c$ and its
native reference for training seed $s$ and reset $r$. For $M$ thrusters with
recorded forces $f_{t,m}$ and opposing-grasp indicator $g_t$, define
\begin{equation}
F=\sqrt{\frac{\sum_{t\in\mathcal{T}}\sum_{m=1}^{M}f_{t,m}^{2}}
 {M|\mathcal{T}|}},\qquad
G=\frac{\sum_{t\in\mathcal{T}}g_t}{|\mathcal{T}|}.
\label{eq:force-grasp}
\end{equation}
Motor-force RMS $F$ measures actuator demand; grasp occupancy $G$ measures the
fraction of sampled time satisfying the opposing-grasp condition. Neither is
conditioned on eventual success. Additional traces measure mechanism progress,
joint tracking, station-position and attitude RMS, force realization and
allocation saturation. A zero-exposure measurement is undefined, not a
zero-effort or zero-contact episode.

Episode summaries are averaged equally over resets and training runs. For
summary $z$, the paired change is
\begin{equation}
\Delta z=\frac{1}{3}\sum_{s\in\{17,43,101\}}\frac{1}{30}
 \sum_{r=1}^{30}(z_{c,s,r}-z_{0,s,r}).
\label{eq:paired-change}
\end{equation}
Completion and grasp differences are expressed in percentage points; force
differences are in newtons. Pairing ensures that a contrast compares the same
initial states and active exposure, rather than two groups selected by their
outcomes. The traces retain both the aggregate summaries and their
checkpoint/reset identities, allowing a reported change to be inspected at
the episode level.

The two horizons answer separate questions. The full episode determines
whether every completion requirement is eventually satisfied. The common
early window permits a like-for-like comparison of interaction and vehicle
response before differing termination times could select different samples.
We therefore keep the final outcome next to the diagnostic summary throughout
the analysis. A grasp fraction calculated over this window cannot certify
an eventual hold or completed rotation, and an early force measurement does
not summarize the duration of the complete task. Reporting exposure makes
the measurement's support visible when a policy fails, stalls or completes
at a different time from its paired reference.

\subsection{Training variation and uncertainty}
We report per-run counts, training-run means and sample SDs. A training run is
the independent unit for training variation; its 30 resets are conditional
evaluations, not additional training replications~\cite{agarwal,henderson}.
Binomial intervals describe reset uncertainty conditional on a
checkpoint~\cite{clopper}. Descriptive crossed bootstrap intervals resample
training runs and reset IDs separately. Training draws are independent across
reference implementations and paired for interventions on a fixed policy;
reset draws are paired throughout. RGB reassignment exchanges images between
15 pairs of batch slots, so its intervals resample those pairs as clusters.
The complete resampling specification is retained in Supplementary
Section~S1.

Intervals are marginal, descriptive summaries of the evaluated training runs
and resets, without simultaneous coverage or a claim of a general ranking.
An interval spanning zero leaves its direction unresolved and does not
establish equivalence. Observed boundary outcomes are given as counts: a
degenerate empirical interval after 0/90 or 90/90 does not remove uncertainty
about future training runs or reset draws. These conventions apply to all
subsequent tables and figures.

\section{Reference Policy Performance}
\begin{figure*}[tbp]
\centering
\includegraphics[width=0.96\textwidth]{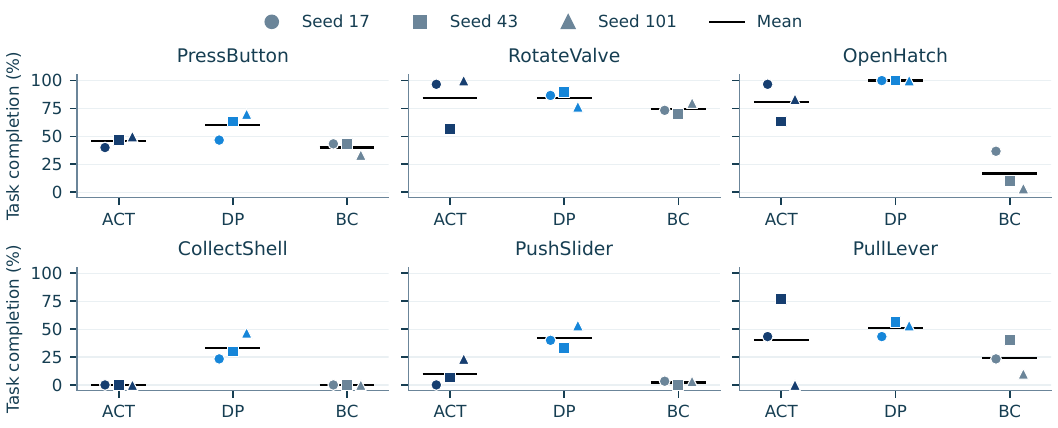}
\caption{ACT, DP and BC on the six tasks. Each marker represents
a separately trained checkpoint on 30 resets; circles, squares and triangles
identify seeds 17, 43 and 101. Black marks show training-run means.}
\label{fig:core-variation}
\end{figure*}

The primary study comprises 54 training runs and 1,620 independently rescored
episodes across six tasks. Table~\ref{tab:core-results} reports the reference
completion rates; Fig.~\ref{fig:core-variation} retains individual training
runs. The purpose of this comparison is to establish native behavior and its
variation before examining what the aggregate outcomes conceal.

\begin{table*}[tbp]
\centering
\caption{Reference completion (\%), mean $\pm$ sample SD over three training
runs and 30 resets per run. ACT, DP and BC cover six tasks; pretrained SmolVLA
covers three, with dashes marking unevaluated tasks. Within each task, all
implementations use the same demonstration split and 320,000-window exposure;
architectures and compute differ.}
\label{tab:core-results}
\small\setlength{\tabcolsep}{7pt}
\begin{tabular}{lrrrr}
\toprule
Task & ACT & DP & BC & SmolVLA \\
\midrule
PressButton & 45.6 $\pm$ 5.1 & 60.0 $\pm$ 12.0 & 40.0 $\pm$ 5.8 & 54.4 $\pm$ 5.1 \\
RotateValve & 84.4 $\pm$ 24.1 & 84.4 $\pm$ 6.9 & 74.4 $\pm$ 5.1 & --- \\
OpenHatch & 81.1 $\pm$ 16.8 & 100.0 $\pm$ 0.0 & 16.7 $\pm$ 17.6 & --- \\
CollectShell & 0.0 $\pm$ 0.0 & 33.3 $\pm$ 12.0 & 0.0 $\pm$ 0.0 & --- \\
PushSlider & 10.0 $\pm$ 12.0 & 42.2 $\pm$ 10.2 & 2.2 $\pm$ 1.9 & 8.9 $\pm$ 6.9 \\
PullLever & 40.0 $\pm$ 38.4 & 51.1 $\pm$ 6.9 & 24.4 $\pm$ 15.0 & 7.8 $\pm$ 5.1 \\

\bottomrule
\end{tabular}
\end{table*}

The six-task macro-averages are 61.9\% for DP, 43.5\% for ACT and 26.3\% for
BC. Variation across training runs changes the interpretation of those
aggregates. On RotateValve, ACT and DP both average 84.4\%, but ACT has a
training-run SD of 24.1 percentage points compared with DP's 6.9. On PullLever,
the ACT checkpoints complete 13, 23 and 0 episodes out of 30. A single
checkpoint would therefore give a substantially different account of the
same implementation's reference behavior.

The largest ACT--DP gaps occur on OpenHatch and PushSlider. ACT-minus-DP is
$-18.9$ and $-32.2$ percentage points, with crossed intervals
$[-38.9,-2.2]$ and $[-55.6,-6.7]$, respectively. Intervals on PressButton and
RotateValve span zero. On CollectShell, ACT and BC complete no episodes,
whereas the three DP checkpoints complete 7, 9 and 14/30. These outcomes
identify the behavior of the evaluated configurations under the stated
training budget.

\noindent\textbf{Pretrained VLA baseline.}
We additionally evaluate SmolVLA~\cite{smolvla} on PressButton, PushSlider and
PullLever with three training seeds and the same 320,000-window demonstration
exposure. Completion is 54.4\%, 8.9\% and 7.8\%, respectively, versus DP's
60.0\%, 42.2\% and 51.1\%. This provides a pretrained VLA reference on a task
subset; architectural and training differences preclude a general comparison
of policy families. The constant instruction tests task adaptation, without
evaluating language grounding. Full configurations and per-seed results are
in Supplementary Section~S3.

OpenHatch is particularly useful for testing a success-only interpretation:
DP completes every primary episode, providing a high native reference rate.
The following studies ask whether this observed completion persists under
specific changes, and whether preserved completion implies preserved physical
execution. Each uses its own matched research cohort rather than reusing the
primary-table score as a control.

\section{What Task Success Conceals}
\label{sec:diagnostics}
The paired protocol distinguishes three questions that a completion rate alone
cannot answer. First, does a failed policy face a scene that a reference
execution can still solve? Second, does increased interaction with a mechanism
advance the full task? Third, does unchanged completion preserve the actuator
demand? We examine these questions using fixed DP checkpoints on OpenHatch
and RotateValve, retaining the physical records when an intervention removes
every completion.

\subsection{Expert-solvable scenes where policies fail}
\label{sec:placement}
\begin{figure*}[tbp]
\centering
\includegraphics[width=0.96\textwidth]{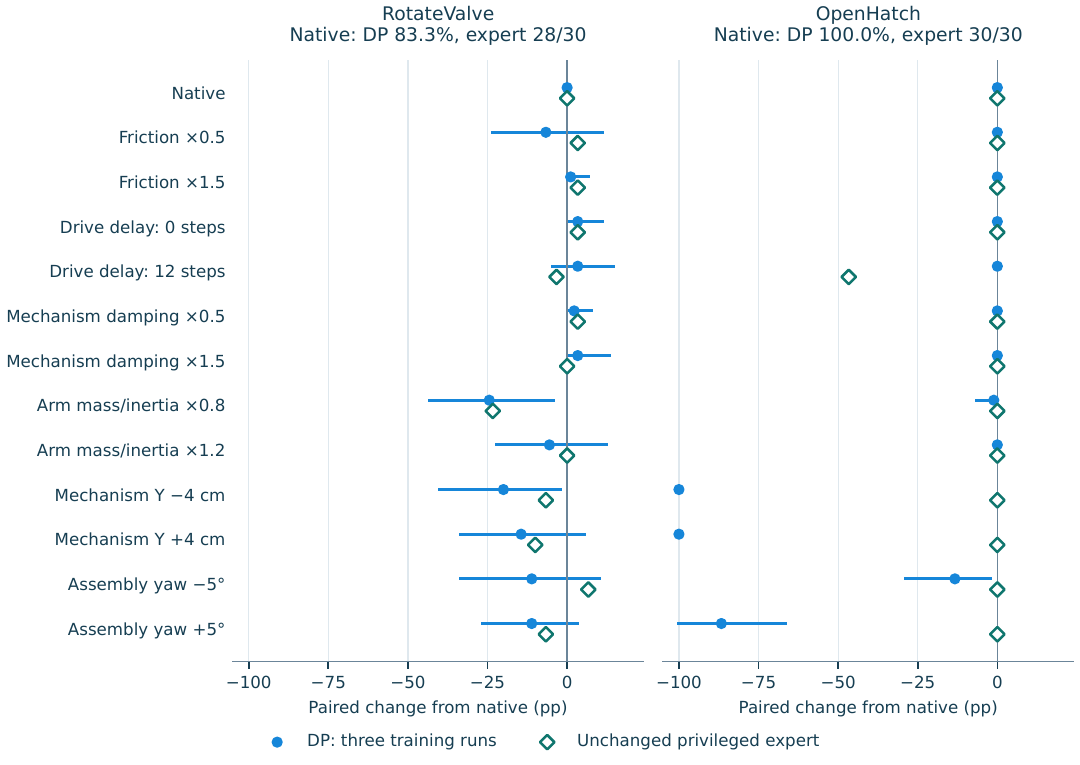}
\caption{Physical interventions on fixed policies. Blue points and bars show
paired DP completion changes and descriptive 95\% crossed intervals over
three checkpoints and 30 resets per checkpoint. Teal diamonds show the
unchanged privileged expert on the same 30 reset identities. Translating
OpenHatch by either $Y$ shift removes every policy completion while preserving
every expert completion.}
\label{fig:physics-sensitivity}
\end{figure*}

Mechanism translation gives the clearest policy--expert separation. The scene
study changes contact friction, actuator delay, mechanism damping, arm mass
and inertia, translation or yaw one factor at a time.
Figure~\ref{fig:physics-sensitivity} compares DP's full-horizon completion with
the unchanged expert on the same reset identities. Each policy condition
contains three training checkpoints and 30 resets per checkpoint; each expert
condition contains the 30 matched reset identities.

For OpenHatch, native DP completion is 100\%. Translating the mechanism and
its supporting geometry by $-4$ or $+4$\,cm along $Y$ reduces completion to
0/90 in each direction. The expert completes 30/30 in both translated scenes.
The expert executions witness solvability under the same physical predicate,
separating the policy's placement sensitivity from scene infeasibility.
Because collection fixes the mechanism pose, these translations also make
the relevant generalization boundary explicit.

Every translated execution retains its native robot, joint and fluid draw.
The unchanged policy and expert use the same reset identities and completion
rules, linking each failure to a completing reference in that scene.

Every evaluated DP checkpoint completes its native hatch cohort and fails
both translated cohorts, so the contrast holds within each training run.
The 90 policy episodes per condition represent three training runs on the
shared 30 reset identities;
the expert's 30 executions witness those identities directly. Resets are
conditional evaluations, rather than independent training replications.

The full traces identify the unmet part of the contract. Neither shifted
condition reaches the required $80$--$105^\circ$ opening window in any of its
90 episodes. At $Y=+4$\,cm, the largest opening over a complete episode remains
below $34^\circ$; at $Y=-4$\,cm it remains below $79^\circ$. A valid final hold
cannot occur before this opening requirement is met. In the positive shift,
early opposing-grasp occupancy nevertheless increases by 10.9 percentage
points (Table~\ref{tab:diagnostic-core}). The conjunction of contact, progress
and full-horizon predicates explains what remains incomplete without assigning
an unobserved fault to a network component. Supplementary
Table~S6 retains the per-condition angle and
hold breakdown.

The extra grasp samples remain contact evidence under the unchanged success
standard. Replaying the same predicate for native, translated and expert
executions distinguishes interaction histories within failed episodes without
promoting contact or partial progress to completion.

Other interventions distinguish policy sensitivity from changes affecting
both policy and reference execution. Yaw shifts of $-5^\circ$ and $+5^\circ$
leave the hatch expert at 30/30 but give DP completion of 86.7\% and 13.3\%.
For RotateValve, reducing arm mass and inertia to $0.8\times$ lowers DP
completion from 83.3\% to 58.9\%, a paired change of $-24.4$ percentage points
with interval $[-43.3,-4.4]$. The expert also declines, from 28 to 21/30.
This second signature indicates that the same physical change affects both
controllers. The expert supplies a different interpretation than in the
translated hatch, rather than a universal measure of task difficulty.

The reference can also perform worse than the policy. With a 12-physics-step
actuator delay, DP preserves every hatch completion, while the expert completes
16/30. Expert success is consequently useful as a concrete witness of
solvability, not as an assumed performance ceiling. The complete factor values,
rates and expert outcomes are retained in Supplementary
Table~S4.

\begin{table*}[tbp]
\centering
\caption{Selected execution signatures from the physical-scene study. Values
are paired changes from each condition's native reference. Completion and
grasp occupancy are in percentage points, attitude RMS in degrees and
motor-force RMS in newtons. Continuous measurements use 600 common active
samples per reset; completion uses the full horizon. Every row averages
three checkpoints and 30 resets each. The full diagnostic matrix is in
Supplementary Table~S5.}
\label{tab:diagnostic-core}
\small\setlength{\tabcolsep}{7pt}
\begin{tabular}{lrrrr}
\toprule
Intervention & Completion & Opposing grasp & Attitude RMS & Motor-force RMS \\
\midrule
Hatch: $Y=+4$ cm & -100.0 & +10.9 & +0.205 & +0.042 \\
Valve: arm $0.8\times$ & -24.4 & -23.6 & +0.517 & +0.746 \\
Hatch: delay 12 & +0.0 & +0.8 & +0.140 & +0.254 \\
\bottomrule
\end{tabular}
\end{table*}

Table~\ref{tab:diagnostic-core} contrasts the execution signatures. The
translated hatch combines a complete loss of completion with increased grasp occupancy and higher
attitude and motor-force RMS. The low-arm-mass valve instead combines reduced
completion with less grasp, larger attitude error and substantially greater
motor-force RMS. Its paired expert also completes fewer resets. Reference
outcomes establish the expert's ability to finish the changed task; execution
measurements describe the policy's physical response.

For the valve arm intervention, attitude RMS changes from $0.543^\circ$ to
$1.060^\circ$, motor-force RMS from 0.841 to 1.587\,N and opposing-grasp
occupancy from 54.5\% to 30.8\%. These recorded changes give physical context
to the completion decrease without requiring a successful policy subset.
Every condition contributes the same active comparison window, including
trials that never finish the mechanism motion. The corresponding contact,
joint-tracking and actuator traces remain inspectable per reset, connecting
the completion decrease to measured stabilization and contact responses.

The delayed hatch supplies a second preserved-completion signature within
this physical-scene cohort. Motor-force RMS increases from 2.645 to
2.899\,N, and requested-versus-realized force RMS increases from 1.578 to
2.314\,N, while full completion is retained. This row illustrates why
actuator measurements remain useful even when the final predicate is
satisfied in every evaluated policy episode. Its baseline belongs to the
delay study; the current experiment below has its own matched native
reference. The release preserves these cohort identities so that similar
physical signatures can be compared without combining their episodes or
substituting one study's baseline for another's.

Matched training ablations provide a narrow follow-up to the placement
failure: neither observation history nor the tested expert trajectory mixture
produces a completion at either $\pm4$\,cm shift. Those designs and full results
are preserved in Supplementary Section~S8. The main diagnostic
claim does not depend on interpreting the architectural or data modification
as a general remedy.

\subsection{Contact improvements without task completion}
\label{sec:integral}
The integral-action intervention changes the execution controller while
holding the learned policy fixed. It evaluates three DP checkpoints on
30 ordered research resets for each task, with identical P/D gains, physical
coefficients, command interfaces and task criteria across integral multipliers
$k_I\in\{1,0.5,0\}$. These resets are separate from the physical-scene
cohort above. Thus the intervention isolates the effect of the specified gain
change within each fixed model and checkpoint.

\begin{figure*}[tbp]
\centering
\includegraphics[width=\textwidth]{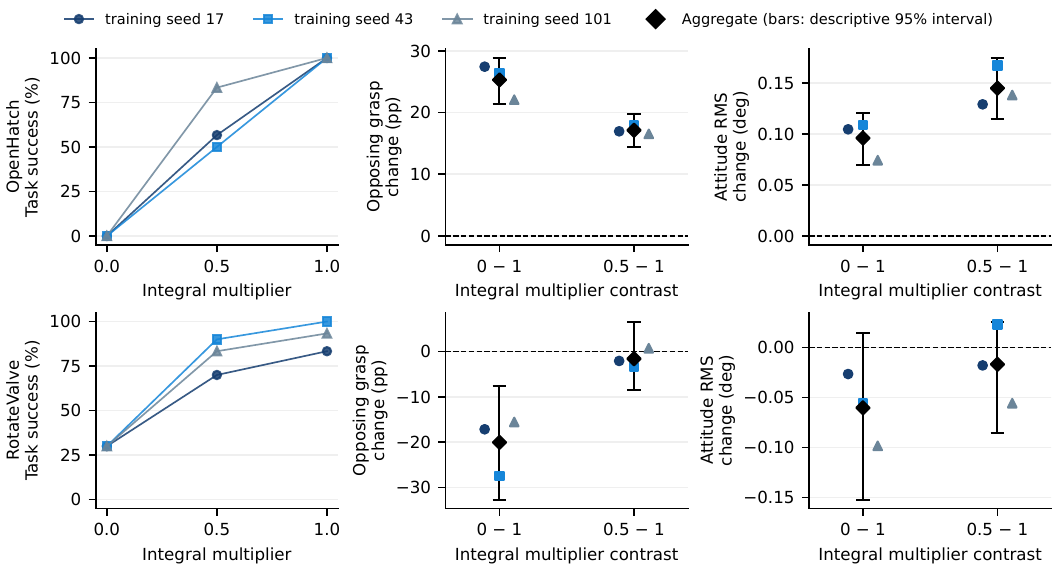}
\caption{Integral action, early interaction and completion. The left panels
show full-horizon success for each fixed DP checkpoint on 30 research resets.
The remaining panels show changes in 20-second opposing-grasp occupancy and
attitude RMS from $k_I=1$. Colored markers identify training runs; black
diamonds and bars give paired means and descriptive 95\% intervals. Removing
integral action increases hatch grasp occupancy while eliminating completion.}
\label{fig:controller-diagnostics}
\end{figure*}

Under nominal coefficients, removing integral action changes OpenHatch
completion from 30/30 to 0/30 for every training seed.
Figure~\ref{fig:controller-diagnostics} compares this full-horizon result with
early contact and stabilization. Opposing-grasp occupancy increases from
3.95\% to 29.25\%, while signed mechanism progress increases from
$2.17^\circ$ to $15.07^\circ$. Every paired execution supplies the complete
20-second diagnostic window. Greater early contact and motion therefore
coexist with failure to complete the unchanged task.

This intervention acts below the high-level policy. Changing the integral
multiplier changes the feedback controller's response while retaining the
other declared gains and the same model. The policy then receives the
resulting observations and can change its subsequent commands through the
ordinary closed loop. Thus the experiment tests the dependence of a fixed
learned implementation on its execution controller, with the native controller
as the paired reference. The known gain intervention supplies the experimental
contrast; the recorded contact, progress and stabilization describe its
physical consequences.

Opposing grasp establishes a contact condition, not successful manipulation
of the mechanism through its complete required motion. Its occupancy can
increase if the robot remains engaged at an unproductive part of an episode.
The measured progress and final predicate are needed to assess the consequence
of that contact. Here the gain change increases early hatch progress without
producing a completing execution. The diagnostic record supports that
observation; it does not turn grasp occupancy into a surrogate objective for
the full contract.

The shared exposure makes the counterexample particularly direct. Failed
executions contribute all of their active samples in the declared window,
and native successes contribute samples over the same duration. The increase
in opposing grasp is therefore not produced by dropping unsuccessful trials
from one condition or by giving the two conditions different observation
lengths. At the same time, an occupancy percentage says how long the opposing
contact condition holds; it does not specify the contact geometry, useful
direction of motion or satisfaction of the final hold. Those distinctions
are resolved by the mechanism and contract traces.

The other task shows that the direction of the contact response is not
shared across mechanisms. RotateValve completion falls from 25, 30 and
28/30 to 9/30 for each seed, a mean change of $-62.2$ percentage points with
interval $[-80.0,-43.3]$. Opposing-grasp occupancy decreases from 57.09\% to
37.05\%, and progress decreases from $95.53^\circ$ to $52.07^\circ$.
Halving the integral gain gives intermediate completion counts on both tasks.
Hatch and valve therefore share the loss of completion but have opposite
early-contact responses.

Vehicle attitude provides another complementary measurement in
Fig.~\ref{fig:controller-diagnostics}. Removing integral action changes hatch
world-identity attitude RMS from $0.472^\circ$ to $0.568^\circ$. On the valve,
it changes from $0.476^\circ$ to $0.416^\circ$, with interval
$[-0.152^\circ,+0.015^\circ]$ for the difference. The latter observed decrease
does not establish a consistent attitude effect, and it accompanies the loss
of valve completion. A stabilization summary therefore needs the manipulation
outcome alongside it, just as a contact summary does. Reading the recorded
axes together retains the difference between how the vehicle moves and
whether the mechanism reaches its required state.

The gain effect persists when base hydrodynamic coefficients are replaced
with sourced alternatives: with the combined damping and added-mass
replacement, removing integral action still gives 0/30 hatch completions for
every seed. The full gain/coefficient matrix and tracking measurements are
in Supplementary Section~S5. They preserve the distinction
between a gain intervention within one model and sensitivity to the model's
coefficients.

\subsection{Unchanged success with increased actuator demand}
\label{sec:current}
The current study asks whether native completion is sufficient to describe
successful execution under disturbance. We apply world-frame currents
$(0,u,0)$ with $u\in\{0,0.10,0.20\}$\,m/s to OpenHatch and RotateValve.
The same three fixed DP checkpoints use 30 fresh research resets per task
across all conditions, including a new zero-current reference, for 540
episodes. Flow begins before camera warmup and acts through robot-link
velocity relative to the water. Gains, coefficients, allocation limits,
rendering and success criteria remain fixed.

\begin{figure}[tbp]
\centering
\includegraphics[width=\columnwidth]{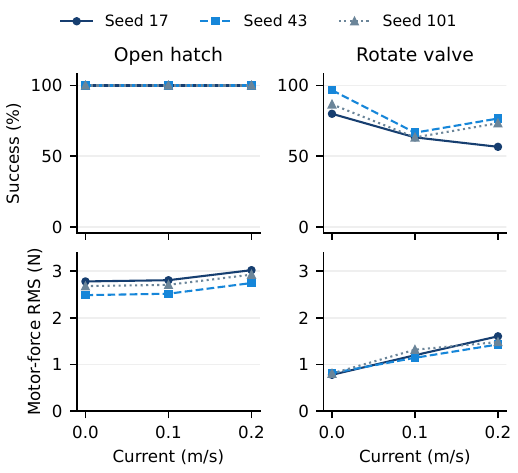}
\caption{Fixed-policy current intervention. Lines identify training seeds;
each condition uses the same 30 resets per checkpoint. Completion uses the
full horizon, while motor-force RMS uses 20\,s. The three hatch completion
curves coincide at 100\%, despite increasing actuator demand.}
\label{fig:current}
\end{figure}

OpenHatch completes 30/30 episodes in every checkpoint and current condition.
Its motor-force RMS nevertheless rises from 2.648\,N at zero current to
2.897\,N at 0.20\,m/s, an increase of 9.4\%
(Fig.~\ref{fig:current}). Every episode supplies the full 20-second comparison
window, and no allocation saturation is recorded. Thus the demand increase
does not rely on selecting successful segments or on comparing exposures of
different lengths. A success-only summary would report the same completion
for all conditions and omit the changed motor-force requirement.

The current contrast preserves the rest of the execution contract. Flow
acts on the robot links through their velocity relative to the water; it
does not directly replace the mechanism dynamics or relax the manipulation
predicate. The increased motor-force RMS is consequently an observed
response of this closed-loop execution to the declared flow condition.
Because allocation limits remain fixed and no early saturation is recorded,
the force trace adds information beyond a binary indicator of whether a
thruster limit was reached. It records a change within the successful
operating conditions of the tested controller.

Completion at its observed ceiling is especially uninformative about this
axis. The hatch satisfies the full predicate at every tested current
magnitude, so the completion data alone give no ordering of actuator demand.
The force measurements distinguish these conditions while preserving their
identical completion result. This is the third diagnostic case, separate
from a failure witnessed by an expert and from a mismatch between contact
and completion. It shows why the same successful outcome can correspond to
different physical execution, even when the policy checkpoint and task are
held constant.

RotateValve supplies a complementary response: completion decreases and the
contact record changes. Mean completion is 87.8\%, 64.4\% and 68.9\% at the
three current magnitudes. Paired changes from zero current are $-23.3$
percentage points ($[-43.3,-1.1]$) and $-18.9$ points ($[-35.6,-2.2]$).
Early contact-loss incidence increases from 1.1\% to 36.7\% and 46.7\%.
Completion and physical execution thus respond differently across the two
mechanisms under the same declared current direction and onset.

The actuator measurement answers a different question from feasibility or
contact. The hatch continues to satisfy its task contract, but the robot
uses a different force profile to do so. That information is useful when
comparing control or policy changes whose completion scores are already near
a boundary. It also remains available for failed episodes on the valve,
allowing the disturbance response to be described across the complete cohort.
Force RMS is interpreted here as motor demand; an energy-consumption claim
would require an additional electrical and temporal model.

\section{Visual Feedback and Sensitivity}
\label{sec:visual}
The physical interventions above change what the robot must execute. Image
interventions instead test whether the information driving that execution
must be current and matched to its physical reset. We evaluate the original
ACT and DP checkpoints on RotateValve and OpenHatch with current RGB,
held-first RGB, or current RGB reassigned between resets. Current measured
proprioception, policy action updates and the original observation-history
implementation are retained in all three modes. A fresh reset block supplies
matched initial physical states and full-horizon evaluation.

\subsection{Image freshness and reset correspondence}
The held-first control repeats each reset's initial wrist frame for the
entire episode. Reassignment exchanges current images between batch slots
15 positions apart in the ordered batch of 30. The latter image changes over
time but comes from the other member of a reset pair. This separates a
time-varying view of the shared task scene from correspondence to the current
robot's own reset.

\begin{figure*}[tbp]
\centering
\includegraphics[width=0.94\textwidth]{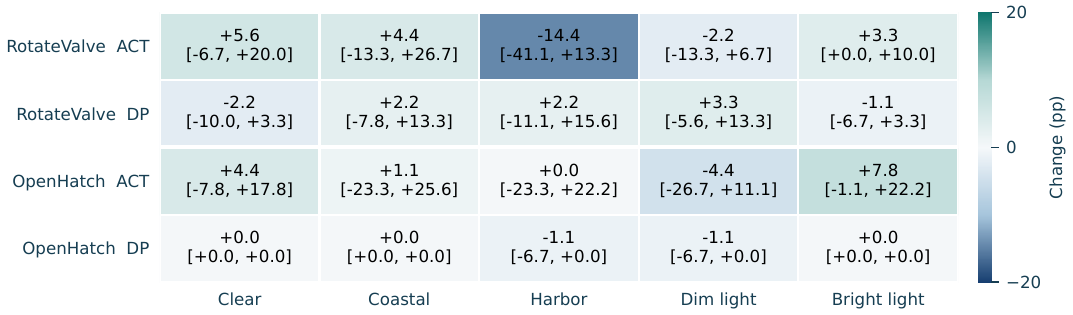}
\caption{Paired completion changes from native appearance and lighting (pp).
Each cell averages three checkpoints and 90 full-horizon episodes. Brackets
give descriptive 95\% crossed intervals. Synthetic water profiles share
native lighting; dim and bright change only scene lighting.}
\label{fig:visual-transfer}
\end{figure*}

\begin{table*}[tbp]
\centering
\caption{RGB interventions. Completion is mean $\pm$ sample SD (\%) over
three checkpoints and 30 resets each; boundary entries show counts out of
90. Paired changes and descriptive 95\% intervals are in percentage points.
Reassigned contrasts resample 15 exchange-pair clusters.}
\label{tab:rgb-controls}
\small\setlength{\tabcolsep}{4pt}
\begin{tabular}{lrrrrr}
\toprule
Task / policy & Current & Held first & Reassigned & Change: held first & Change: reassigned \\
\midrule
RotateValve / ACT & $85.6\pm22.2$ & 0/90 & 0/90 & $-85.6\;[-100.0,-60.0]$ & $-85.6\;[-100.0,-60.0]$ \\
RotateValve / DP & $77.8\pm19.5$ & 0/90 & 0/90 & $-77.8\;[-100.0,-55.6]$ & $-77.8\;[-100.0,-56.7]$ \\
OpenHatch / ACT & $84.4\pm21.2$ & 0/90 & $88.9\pm7.7$ & $-84.4\;[-100.0,-60.0]$ & $+4.4\;[-13.3,+23.3]$ \\
OpenHatch / DP & 90/90 & 0/90 & 90/90 & $-100.0\;[-100.0,-100.0]$ & $+0.0\;[+0.0,+0.0]$ \\

\bottomrule
\end{tabular}
\end{table*}

Holding the first image removes every completion for both tasks and both
implementations (Table~\ref{tab:rgb-controls}). Current robot state and
continued action updates do not preserve the reference behavior when the
image remains fixed. The valve DP trace gives a physical account of this
outcome: opposing-grasp occupancy falls from 48.5\% to zero, and the mean
episode maximum rotation in the shared early exposure falls from
$86.1^\circ$ to $1.4^\circ$. Contact and mechanism motion change alongside
the final completion outcome.

Reassignment has a different effect. It eliminates every valve completion,
whereas hatch ACT completes 88.9\% and hatch DP completes 90/90. The reassigned
frames still contain an evolving scene from the simultaneous batch, while
the held-first frame does not change. Freshness and correspondence therefore
produce different responses and should not be treated as one generic
``vision removed'' condition. The controls describe the tested policies under
these image replacements; a separately trained proprioceptive policy would
be a different comparison.

\subsection{Transfer across synthetic appearance conditions}
We also vary water appearance and lighting while holding physical settings,
policy weights and reset identities fixed. Clear, coastal and harbor presets
apply distance-dependent attenuation and backscatter to wrist RGB using
camera range. Dim and bright conditions change scene lights under native
appearance; particle and refractive settings remain fixed. The conditions
and intervals are shown compactly in Fig.~\ref{fig:visual-transfer}; full
completion rates and the input-image examples remain in Supplementary
Section~S7.

The largest observed decrease is ACT on RotateValve under harbor appearance:
completion changes from 86.7\% to 72.2\%, or $-14.4$ percentage points, with
interval $[-41.1,+13.3]$. DP on the same task changes from 85.6\% to 87.8\%.
For OpenHatch, ACT remains at 87.8\%, and DP changes from 100\% to 98.9\%.
The wide valve ACT interval leaves the direction unresolved across the
sampled runs and resets. These modest appearance responses contrast with
the loss of every completion under held-first RGB. The experiments therefore
separate transfer to the specified appearance transformations from dependence
on changing visual feedback.

\FloatBarrier
\section{Discussion and Limitations}
\label{sec:discussion}
The three diagnostic cases change the interpretation of task success in
different ways. Paired expert execution shows that the shifted hatch remains
solvable despite policy failure. The integral intervention shows that more
early opposing grasp and motion do not ensure completion. Current shows that
successful manipulation can require greater motor-force demand. A single
completion rate cannot distinguish these outcomes, because each comparison
requires evidence about a different part of the coupled execution.

The protocol makes those measurements comparable by fixing the task predicate,
retaining native references for each study and matching both reset states and
active exposure. It preserves continuous evidence when no policy episode
completes. This is especially important near completion boundaries: a zero
rate still has an interpretable contact and progress record, while a full
rate can coexist with changed actuator demand. The expert reference and
execution traces serve complementary roles. One supplies a completing
execution in the changed scene; the other records how the policy behaves
there.

These findings suggest a practical evaluation sequence. Report full-horizon
completion and training-run variation first. For an intervention, compare
each policy against its own native execution on matched resets. Where a
physical scene change causes failure, add an unchanged expert reference on
the same states. Finally, retain contact, progress, tracking and actuator
measurements for every episode over a declared exposure. The result is an
account of observed execution, rather than a diagnosis inferred solely from
the final binary score.

The visual controls extend this reasoning to policy input. Holding the first
image and exchanging current images differ on the hatch, so changing images
and reset-specific correspondence should be tested separately. A policy can
be relatively insensitive to the tested synthetic appearance changes yet
lose all completion under stale images. The adaptive follow-up at
$\pm0.5$ and $\pm1$\,cm describes local placement sensitivity of the existing
history/trajectory-mixture variants; its full results and adaptive status are
retained in Supplementary Section~S8, rather than used as an
independent confirmation of the original placement contrast.

\subsection{Scope and limitations}
The evidence concerns the tested simulation and reference implementations.
Hydrodynamics are reduced-order, optical presets are not validated against
real underwater images, and transfer to hardware remains untested. Learning
uses procedural visuals and box fingers; the separate CAD-profile expert and
platform studies do not establish geometry or platform transfer of learned
policies. The broad reference comparison uses three training runs per
configuration, differing encoders, histories, command representations and
compute, without a systematic hyperparameter search.

Detailed interventions concentrate on RotateValve and OpenHatch. The
supplementary stale-chunk and reduced-capacity tests use one checkpoint per
task. Separate reset blocks and checkpoint subsets define the scope of each
contrast and should not be pooled as one estimate of generality. The expert
is a particular privileged controller, so its failures cannot rule out other
completing executions. Motor-force RMS measures demand, not electrical energy.
Broader mechanisms, additional training runs and physical robot trials are
needed to assess the protocol's generality and the simulation's predictive
value.

\subsection{Reproducibility and conclusion}
The release links task contracts, whole-episode demonstration splits,
training settings, final checkpoints and per-episode physical traces. The
complete expert candidate streams include unsuccessful attempts, making the
selected training data inspectable. Checkpoint, initialization, cohort order,
rollout-noise, precision and source-version records bind reported outcomes to
their executions. Independent physical rescoring and recorded hashes connect
the summaries to that evidence. Code, documentation and datasets are linked
from the \href{https://dancher00.github.io/wasserman/}{project website};
Supplementary Section~S11 gives the full release and reproduction
specification.

Together, the benchmark and paired protocol support standardized policy
learning and evaluation with inspectable physical evidence. The three central
cases demonstrate why completion should be read alongside expert solvability,
contact and progress, and actuator demand. The three-task SmolVLA evaluation
extends the learning references to a pretrained VLA. Additional platforms,
interfaces and training choices remain available as supporting studies.

\begingroup
\makeatletter
\renewcommand\section{\@startsection{section}{1}{\z@}%
  {10\p@ \@plus 3\p@ \@minus 2\p@}%
  {5\p@ \@plus 2\p@ \@minus 1\p@}%
  {\normalsize\bfseries\boldmath}}
\makeatother
\section*{CRediT authorship contribution statement}
\textbf{Danil Belov:} Methodology, Software, Investigation, Formal analysis,
Visualization, Writing -- original draft, Writing -- review \& editing.
\textbf{Artem Erkhov:} Formal analysis.
\textbf{Sergei Parsegov:} Conceptualization.
\textbf{Pavel Osinenko:} Supervision.

\section*{Funding}
No external funding was received for this work.

\section*{Declaration of competing interest}
The authors declare no competing interests.

\section*{Declaration of generative AI and AI-assisted technologies
in the manuscript preparation process}
ChatGPT assisted with manuscript formatting and presentation. OpenAI image
generation produced the explanatory robot and visual-effect icons in Figure~\ref{fig:teaser}.
The authors checked sources and outputs and take full responsibility for
the manuscript.

\endgroup
\clearpage
\balance

\end{document}